\documentclass[runningheads]{llncs}
\usepackage{graphicx}
\usepackage{multicol}
\usepackage{multirow}
\usepackage{rotating}
\usepackage{amsmath}

\usepackage{amsmath}
\usepackage{xspace}
\usepackage[T1]{fontenc}
\usepackage{booktabs}

\usepackage{graphicx}
\usepackage{amsmath,amssymb}
\usepackage{float}
\begin{document}
\title{ Inference-Time Orthogonal Seeding Enables Geometry-Aligned 3D Organ Segmentation for Slice-Propagation Methods}
\titlerunning{Orthogonal Seeded Slice-Propagation}

\author{Md Rakibul Haque\inst{1,2,3} \and
Tushar Kataria\inst{1,2} \and 
Shireen Y. Elhabian \inst{1,2} 
\authorrunning{F. Author et al.}
\institute{Kahlert School of Computing, University of Utah, Salt Lake City, USA \and
Scientific Computing and Imaging Institute, University of Utah, Salt Lake City, USA \and  Corresponding author\\ \
{\{rakibul,tushar.kataria,shireen\}@sci.utah.edu }}
}
\maketitle              % typeset the header of the contribution
\begin{abstract}
Dense voxel-level annotation remains a major bottleneck in 3D medical image segmentation. Single-slice propagation methods such as Sli2Vol reduce this burden by propagating one annotated seed slice through a volume using label-free registration. However, axial-only propagation accumulates errors with distance from the seed, especially in surface-distance metrics, because it ignores coronal and sagittal evidence and therefore underuses the 3D information available in CT/MRI volumes.
To better leverage volumetric geometry, we study how key training and inference choices affect slice-propagation models, including single-axis versus multi-axis label-free registration, single-seed versus multi-seed propagation, and orthogonal seed configurations. Instead of propagating from a single axial seed, we use three orthogonal seeds—one axial, one coronal, and one sagittal—and fuse their propagated labels with a simple label-free rule.
Our results show that the training paradigm has limited impact: an axially trained network applied to off-axis seeds captures nearly all the improvement, while explicit three-axis training adds little. Instead, performance is driven by inference-time seed geometry, especially orthogonality rather than the number of annotated slices, as a budget-matched three-axial control provides no benefit and can even degrade performance. On a multi-organ CT cohort, orthogonal seeding with the axial Sli2Vol backbone improves Dice by 21.9\%, Normalized Surface Dice by 25.5\%, and reduces Average Hausdorff Distance by 53.5\% over the single-axis baseline.The code is publicly available at:
\url{https://github.com/RakibulHaqueSajal/SlicePropagation}.

\end{abstract}

\section{Introduction}

%Volumetric segmentation underpins diagnosis, treatment planning, and quantitative analysis in 3D CT and MRI , yet dense voxel-level annotation requires an expert to delineate each organ across tens to hundreds of slices, scaling with every structure, modality, and protocol. Fully supervised methods such as nnU-Net \cite{isensee2021nnunet} depend on large curated label sets, while promptable models such as SAM \cite{kirillov2023sam} and MedSAM \cite{ma2024medsam} operate slice by slice and need repeated interaction to cover a volume. These costs motivate \emph{semi-automatic} propagation, where a user annotates one or a few slices and an algorithm fills the rest. A prominent instance is Sli2Vol \cite{yeung2021sli2vol}, a label-free inter-slice registration network that learns by reconstructing one slice from its neighbor \cite{wang2019timecycle} and then transports a seed mask through the volume. Such propagation can be realized in one of two ways, either by matching features between slices, the \emph{correspondence-based} paradigm of Sli2Vol, or by predicting a dense deformation field, the \emph{deformable} paradigm of Vol2Flow \cite{bitarafan2022vol2flow}, Flow2Mask \cite{bitarafan2025flow2mask}, VoxelMorph \cite{balakrishnan2019voxelmorph}, and TransMorph \cite{chen2022transmorph}.

Volumetric segmentation is central to diagnosis, treatment planning, and quantitative analysis in 3D CT and MRI. Fully supervised methods such as nnU-Net \cite{isensee2021nnunet} rely on large curated label sets; however, dense voxel-level annotation requires experts to delineate each organ across tens to hundreds of slices, with effort scaling across structures, modalities, and imaging protocols. Promptable foundation models such as SAM \cite{kirillov2023sam} and MedSAM \cite{ma2024medsam} operate slice by slice, but still require repeated interaction to segment a full volume. These annotation costs motivate \emph{semi-automatic} propagation methods, where a user annotates one or a few slices and an algorithm propagates the labels through the remaining volume. A representative example is Sli2Vol \cite{yeung2021sli2vol}, a label-free inter-slice registration framework that learns slice-to-slice correspondence by reconstructing one slice from its neighbor \cite{wang2019timecycle}, as a meta-training task, and uses the learned correspondence to propagate a seed mask through the volume. More broadly, slice propagation can follow either a \emph{correspondence-based} paradigm, where masks are transported using learned feature matching, as in Sli2Vol, or a \emph{deformable} registration paradigm, where a dense deformation field is estimated, as in Vol2Flow \cite{bitarafan2022vol2flow}, Flow2Mask \cite{bitarafan2025flow2mask}, VoxelMorph \cite{balakrishnan2019voxelmorph}, and TransMorph \cite{chen2022transmorph}.

These methods share structural limitations that remain underexplored. First, propagation is typically restricted to a single axis, usually the axial ordering, leaving coronal and sagittal planes unused even though boundaries that are ambiguous in one orientation may be clearer in another. The complementarity of multi-orientation views is well established: 2.5D and triplanar networks fuse axial, coronal, and sagittal evidence for detection, registration and segmentation \cite{prasoon2013triplanar,setio2016multiview,sundaresan2021truenet}, and multi-atlas segmentation has long combined multiple label estimates using spatially varying, distance-aware weighting rules \cite{artaechevarria2009combination,bin2025efficientmorph}, while orthogonal annotations have also been combined with registration-based propagation for weakly supervised segmentation \cite{cai2023orthogonal}. However, their direct use as inference-time seeds in existing slice-propagation pipelines remains less explored . Second, prediction proceeds as a sequential chain from a single seed slice, causing errors to accumulate and accuracy to degrade with distance from the seed, most visibly in surface-based metrics\cite{nihalaani2024estimation}. Together, these limitations prevent slice-propagation methods from fully using the 3D geometry available in volumetric data, raising two central questions: is propagation quality determined mainly by the registration model and its training, or by how seed annotations are geometrically configured and used at inference? More importantly, can alternative seeding strategies substantially improve segmentation accuracy while requiring only a modest increase in manual annotation effort?

To answer these questions, we conduct a controlled study that decouples the registration/correspondence model and its training from the inference-time seed configuration. We compare three propagation backbones: an axial Sli2Vol model trained only for axial slice correspondence, a three-axis Sli2Vol model trained to register slices along the axial, coronal, and sagittal directions, and a deformable TransMorph model that predicts dense deformation fields instead of correspondence maps. These experiments isolate the effect of different label-free meta-training strategies on downstream organ segmentation. In parallel, we move beyond single-seed propagation and evaluate different inference-time seed configurations, including additional seeds along the same axis and orthogonal seeds provided along axial, coronal, and sagittal planes. This allows us to determine whether segmentation accuracy is driven primarily by the learned registration model or by the geometry of the annotated seeds used at inference. The paper makes the following contributions:
\begin{itemize}
\item We study how the label-free meta-training task, correspondence versus registration, and the inference-time seed configuration, single-slice, multi-slice, or orthogonal-slice, affect slice-propagation models.
\item Our results show that orthogonal seeding produces substantially better 3D segmentation at the cost of only two additional annotations.
\end{itemize}
\vspace{-2em}
\section{Method}

\begin{figure}[t]
\centering
\includegraphics[width=\textwidth]{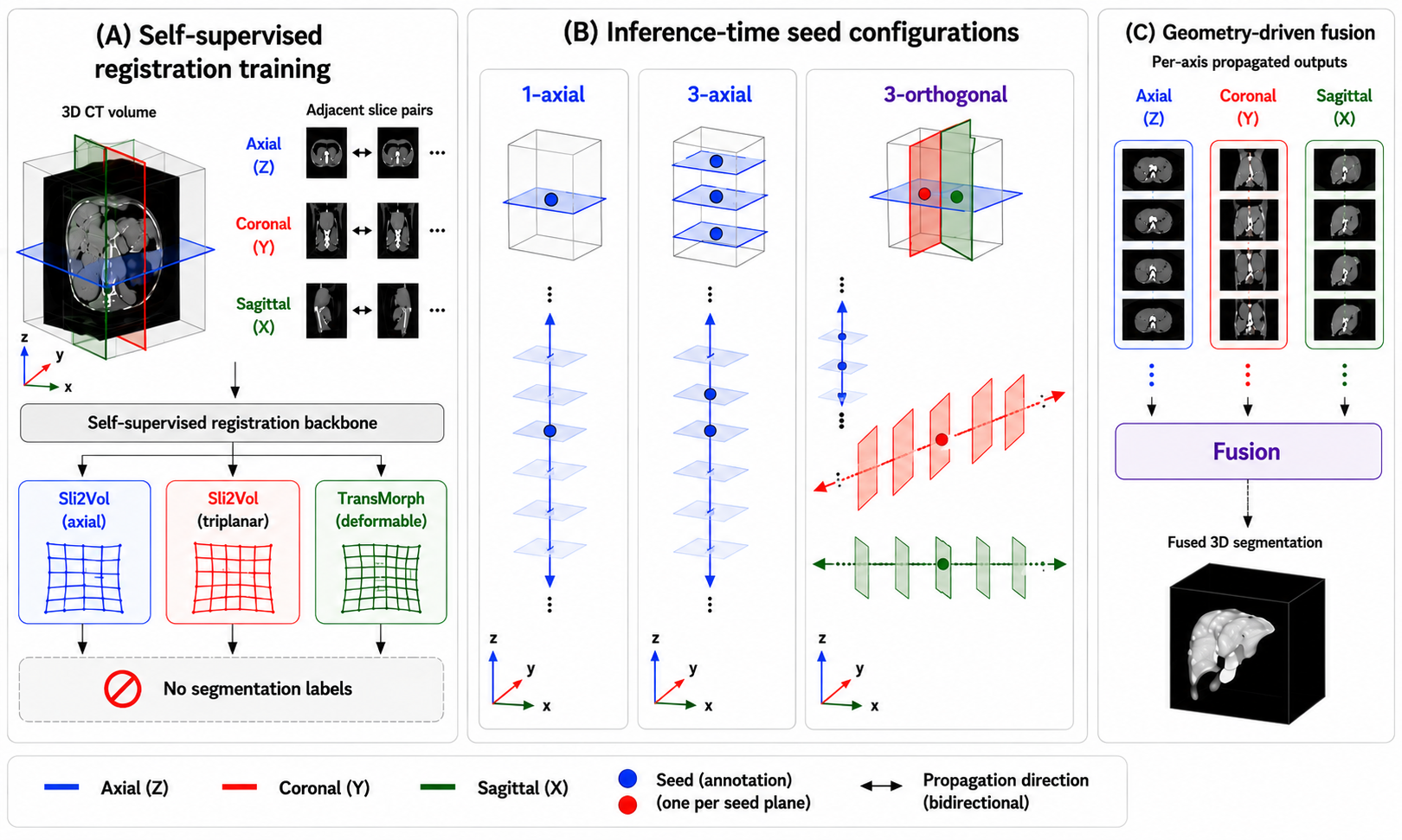}
\caption{Overview. (A) A self-supervised registration backbone is trained on adjacent slice pairs with no labels, giving axial Sli2Vol, triplanar Sli2Vol, and deformable TransMorph. (B) At inference we compare one axial seed, three coplanar axial seeds, and three orthogonal seeds. (C) The per-axis propagated maps are combined by a label-free distance-weighted fusion into a 3D segmentation.}
\label{fig:method}
\end{figure}

We decompose single-slice mask propagation into two independent design choices: (a) the registration backbone \& its training, and (b) the seed configuration used at inference (Fig.~\ref{fig:method}). By varying one factor while holding the other fixed, we isolate the source of downstream segmentation performance.

\textbf{Single Slice Propagation:}
Let $\mathbf{V}\in\mathbb{R}^{H\times W\times Z}$ be a CT volume with slice $V_k$ along a chosen axis. A correspondence backbone defines a transport operator $\mathcal{T}(V_a,V_{a\pm 1})$ that maps a mask on slice $V_a$ onto the neighbor $V_{a\pm 1}$. Given a seed slice $s$ with user mask $M_s$, the mask is propagated bidirectionally from the seed,
\begin{equation}
\tilde{S}_{k\pm1}=\mathcal{T}(V_k,V_{k\pm1})\big[\tilde{S}_k\big],\qquad \tilde{S}_s=M_s,
\end{equation}
which yields a full-volume soft prediction $p\in[0,1]^{H\times W\times Z}$ and its binarization $\mathbf{P}=\mathbb{1}[p>0.5]$. Every backbone is trained self-supervised, reconstructing one slice from its neighbor from image slices alone. We test two paradigms below that differ only in how the operator $\mathcal{T}$ is realized: by matching features, or by predicting a deformation field.
\par

\textbf{\textit{Correspondence-based (Sli2Vol).}}
The first paradigm realizes $\mathcal{T}$ by feature matching. Sli2Vol~\cite{yeung2021sli2vol} forms a local affinity matrix $\mathbf{A}_{a\to b}$ between a slice pair, so transport is a weighted copy $\mathcal{T}(V_a,V_b)[X]=\mathbf{A}_{a\to b}X$, trained by the label-free slice-reconstruction loss
\begin{equation}
\mathcal{L}_{\mathrm{rec}} = \big\|\,\mathbf{A}_{a\to b}\,V_a - V_b\,\big\|_1 .
\end{equation}
We compare two training regimes that differ only in adjacent-pair sampling. The \emph{axial} regime samples slice pairs only along the axial axis, as in the original Sli2Vol, whereas the \emph{triplanar} regime samples pairs from the axial, coronal, and sagittal axes, exposing the same network to all three orientations without changing the architecture or objective. Both regimes use the original Sli2Vol hyperparameters.

\par
\textbf{\textit{Deformable (TransMorph).}}
The second paradigm instead realizes $\mathcal{T}$ by an explicit warp. TransMorph~\cite{chen2022transmorph} predicts a deformation field $\phi_{a\to b}=\Phi_\theta(V_a,V_b)$ and warps through a spatial transformer, $\mathcal{T}(V_a,V_b)[X]=X\circ\phi_{a\to b}$, trained by a local normalized cross-correlation term with a diffusion regularizer,
\begin{equation}
\mathcal{L}=\mathcal{L}_{\mathrm{LNCC}}\big(V_a\circ\phi_{a\to b},\,V_b\big)+\lambda\,\big\|\nabla\phi_{a\to b}\big\|_2^2 .
\end{equation}
We train the deformable model on all three axes, mirroring the triplanar regime, so both paradigms are exposed to the same multi-axis data and any inference-time gain cannot be attributed to orientation coverage during training. Because the model outputs a hard-warped mask rather than soft probabilities, the fusion step below is restricted to hard labels.
\par 

\textbf{Inference Time Seed configurations.}
Given any trained backbone we compare three configurations (Fig.~\ref{fig:method}B). Unless stated otherwise, the seed on axis $a$ is the slice of largest target cross-sectional area,
\begin{equation}
s_a=\operatorname*{arg\,max}_i \big|\mathbf{M}^{(a)}_i\big|,
\end{equation}
a standardized proxy for an informative user-selected slice.For controlled evaluation, the ground-truth mask is used only to identify this seed slice; practical user-driven slice selection and its associated effort are not evaluated here. \emph{\textbf{1-axial}} propagates a single axial seed, the standard setting. \emph{\textbf{3-axial}} propagates three axial seeds at $25/50/75\%$ of the organ's extent and fuses them, a budget-matched control that spends the same annotation as the orthogonal setting but keeps the seeds coplanar. \emph{\textbf{3-orth}} propagates one max-area seed per orientation along its own axis, maps each back to the canonical $(H,W,Z)$ frame, and fuses.

\textbf{Distance-weighted fusion.}
Each chain $c$ is most reliable near its own seed, so we weight it at voxel $v$ by an exponential of the distance along its axis $a(c)$ to the seed $s_c$ (Fig.~\ref{fig:method}C):
\begin{equation}
\omega_c(v)=\exp\!\big(-\big|[v]_{a(c)}-s_c\big|/\sigma\big),
\qquad
p_{\mathrm{fuse}}(v)=\frac{\sum_c \omega_c(v)\,p_c(v)}{\sum_c \omega_c(v)},
\end{equation}
with $\hat{y}(v)=\mathbb{1}[p_{\mathrm{fuse}}(v)>0.5]$. The weight is exogenous, depending only on seed geometry and not on predictions, so no labeled data tunes the fusion; this is also why orthogonal seeds help, since three axes shorten the effective distance to some seed for any voxel. We use $\sigma=30$ slices, stable for $\sigma\in[20,50]$. Because TransMorph gives no soft map, its chains are fused by label-free majority vote; for the correspondence backbone we also report equal-weight averaging ($\omega_c\equiv1$) as an ablation isolating the distance weighting.
\section{Results}
\vspace{-0.5em}
\textbf{Datasets.}
We train and evaluate on publicly available CT collections with disjoint cohorts for each stage. Self-supervised training uses five unlabeled collections: CHAOS~\cite{kavur2021chaos} (40 volumes), MSD Liver~\cite{simpson2019msd} (70), Pancreas-CT~\cite{roth2015pancreas} (80), KiTS~\cite{heller2019kits} (210), and CT Lymph Nodes~\cite{roth2014lymph} (176); no segmentation labels are used. Evaluation covers three structures: SLIVER07~\cite{vanginneken2007sliver} (20 liver CT volumes), Decathlon-Pancreas~\cite{simpson2019msd} (281), and Decathlon-Spleen~\cite{simpson2019msd} (41). No training set shares patients with any test set, and spleen is unseen during training, making spleen performance a test of structure-agnostic generalization.

\textbf{Implementation Details.}
Our correspondence backbone uses the original Sli2Vol~\cite{yeung2021sli2vol} Sli2Vol-edge setting: edge radius $3$, $R=6$, $\tau=0.7$, and $\Delta z=1$. We train \emph{Axial} and \emph{Triplanar} variants, where triplanar sampling draws one adjacent pair per axis per volume each epoch. Both use Adam ($\text{lr}=10^{-4}$, batch size $16$) for up to $350$ epochs with early stopping patience $50$. As a deformable counterpart, we train 2D TransMorph~\cite{chen2022transmorph} on the same corpus with local normalized cross-correlation loss and diffusion regularization ($\text{lr}=10^{-4}$, batch size $8$). All models are trained self-supervised without segmentation labels.

Volumes are not resampled to isotropic spacing. Orthogonal views are obtained by axis permutation of the stored $(Z,H,W)$ volume, with each slice resized to $256\times256$. Distance weights are computed in slice counts, while surface metrics use native voxel spacing. At inference, each axis uses the slice with the largest target cross-section as the seed, and distance-weighted soft fusion uses fixed $\sigma=30$ slices without label-based tuning.

Orthogonal seeding changes only inference, propagating one chain per axis instead of one. This gives an analytical $\sim!3\times$ increase in forward passes and, on a single GPU, roughly $45$,s per volume versus $27$,s for the single-axis baseline, with negligible fusion overhead. We report Dice ,  Normalized surface dice (NSD) at $2,\text{mm}$, and Average Hausdorff Distance (AHD, mm), averaged per cohort against the single-axis Sli2Vol baseline. Significance is tested with a two-sided paired Wilcoxon signed-rank test on per-case Dice ($n{=}342$).

\textbf{Quantitative Results.}
Table~\ref{tab:main} evaluates each backbone under all inference configurations on the multi-organ test cohort, crossing registration model and training strategy against inference-time seed configuration.

\begin{table}[!htbp]
\centering
\caption{Segmentation quality across three backbones and three seed configurations, as mean$\pm$std over the test set. Dice and NSD@2mm are higher-better; AHD (mm) is lower-better. The orthogonal configuration (\textbf{bold}) is best for every backbone on Dice and AHD, and for the correspondence backbones on NSD.}
\label{tab:main}
\setlength{\tabcolsep}{4pt}
\renewcommand{\arraystretch}{1.15}
\resizebox{\textwidth}{!}{%
\begin{tabular}{|l|l|c|c|c|}
\hline
Backbone & Metric & 1-axial & 3-axial & 3-orthogonal \\
\hline\hline
\multirow{3}{*}{Sli2Vol (axial)}
 & Dice     & $0.595\pm0.170$ & $0.542\pm0.217$ & $\mathbf{0.725\pm0.134}$ \\
 & NSD@2mm  & $0.469\pm0.165$ & $0.393\pm0.194$ & $\mathbf{0.589\pm0.141}$ \\
 & AHD (mm) & $7.28\pm3.49$   & $10.76\pm6.48$  & $\mathbf{3.39\pm1.86}$   \\
\hline
\multirow{3}{*}{Sli2Vol (triplanar)}
 & Dice     & $0.614\pm0.169$ & $0.558\pm0.222$ & $\mathbf{0.745\pm0.135}$ \\
 & NSD@2mm  & $0.512\pm0.161$ & $0.404\pm0.211$ & $\mathbf{0.617\pm0.151}$ \\
 & AHD (mm) & $6.58\pm3.37$   & $10.41\pm6.41$  & $\mathbf{2.88\pm1.71}$   \\
\hline
\multirow{3}{*}{TransMorph (deformable)}
 & Dice     & $0.575\pm0.174$ & $0.321\pm0.148$ & $\mathbf{0.661\pm0.089}$ \\
 & NSD@2mm  & $\mathbf{0.474\pm0.163}$ & $0.153\pm0.086$ & $0.424\pm0.097$ \\
 & AHD (mm) & $7.60\pm3.67$   & $34.10\pm15.38$ & $\mathbf{6.48\pm2.36}$   \\
\hline
\end{tabular}%
}
\vspace{-2em}
\end{table}

Two patterns emerge from Table~\ref{tab:main}. First, within a fixed seed configuration, changing the backbone or training strategy has only a modest effect. The axial and triplanar correspondence models differ only in training, yet remain within $0.02$ Dice under orthogonal seeding, while the deformable backbone has lower absolute performance but follows the same ordering. Second, within each backbone, the seed configuration is the dominant factor. Orthogonal seeding achieves the best Dice and AHD for every backbone, whereas the budget-matched three-axial control uses the same number of annotations but does not improve over a single axial seed.

This indicates that the gain comes from seed orthogonality rather than annotation count. For TransMorph, the three-axial control collapses in AHD ($7.60\to34.10$,mm), likely because coplanar seeds produce correlated errors and hard majority voting amplifies shared drift instead of canceling it. The same hard-vote limitation explains why orthogonal seeding does not improve performance for TransMorph NSD ($0.474\to0.424$): noisy off-axis boundaries cannot be down-weighted as in soft correspondence fusion. Thus, this exception reflects the fusion mechanism rather than a failure of orthogonal seeding.

For axial Sli2Vol, orthogonal seeding improves cohort-mean Dice by $21.9\%$, but the gains are even more pronounced on surface-sensitive metrics: NSD increases by $25.5\%$ and AHD is reduced by $53.5\%$ over the single-axis baseline. This indicates that orthogonal seeding not only improves volumetric overlap but also substantially corrects boundary drift and long-range propagation errors, with the largest Dice gain on pancreas ($+28\%$). All orthogonal-versus-baseline Dice improvements are highly significant under paired Wilcoxon signed-rank tests: axial $3$-orthogonal versus $1$-axial ($p<0.001$), triplanar $3$-orthogonal versus $1$-axial ($p <0.001$), and $3$-orthogonal versus the budget-matched $3$-axial control ($p<0.001$), confirming that the improvement is geometry-driven rather than annotation-count-driven.

\begin{figure}[!b]
\centering
\includegraphics[width=\textwidth]{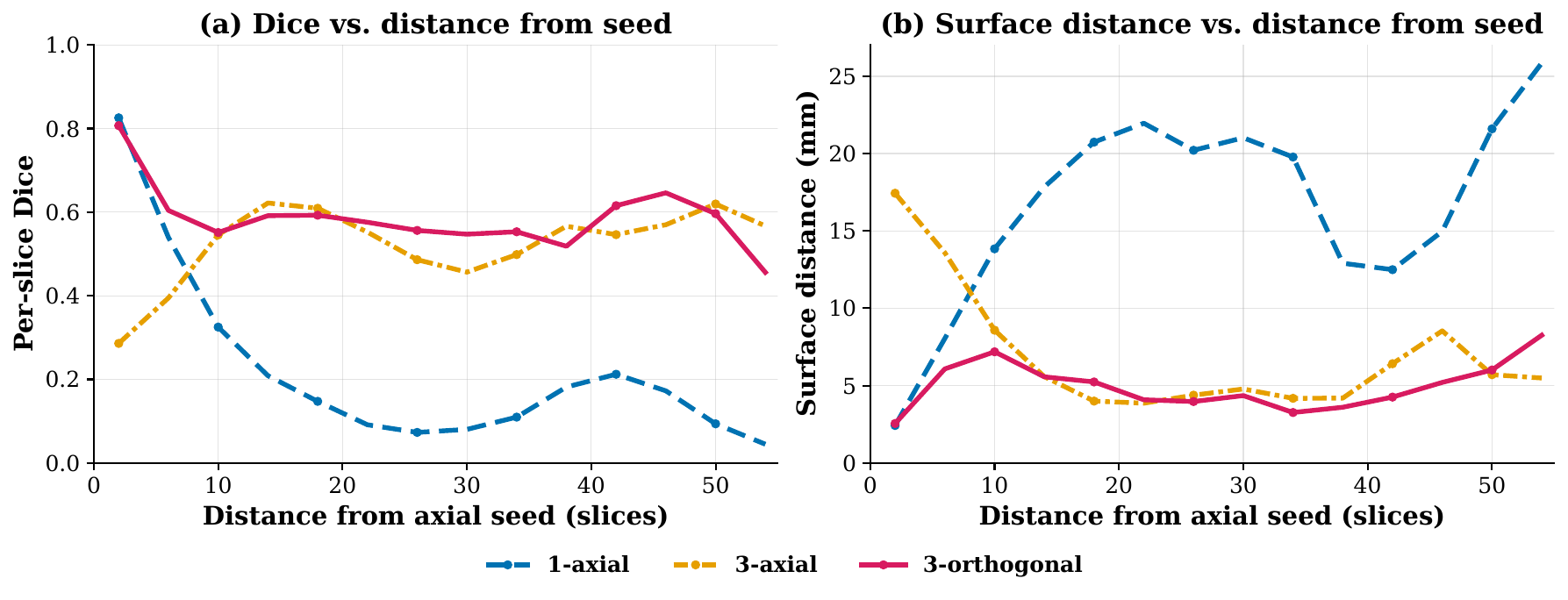}
\caption{Segmentation quality versus distance to the nearest seed, on the triplanar Sli2Vol backbone. Left: per-slice Dice. Right: per-slice surface distance (mm). The single-axis baseline (1-axial) decays away from its lone seed, the coplanar control (3-axial) fills the mid-range but starts poorly near the shared seed plane, and the orthogonal configuration (3-orthogonal) stays comparatively high and flat throughout. All curves come from one trained network, so the differences reflect seed geometry alone.}
\label{fig:distcurves}
\end{figure}

\textbf{Propagation Error Analysis}. Figure~\ref{fig:distcurves} makes the mechanism propagation error accumulation explicit. Since propagation is sequential, each slice's accuracy depends on its distance to the nearest reliable seed. With 1-axial propagation, Dice is highest near the seed but decays toward the axial poles, while surface distance increases, illustrating the core failure mode of single-slice propagation. Adding two coplanar seeds in the 3-axial setting improves the mid-range but introduces disagreement near the shared seed plane, where Dice drops to $\approx0.29$ and surface distance rises to $\approx17$,mm in the nearest bin, and performance still degrades near the organ extremes. In contrast, 3-orthogonal seeding places seeds on mutually perpendicular planes, so slices far from the axial seed remain close to a coronal or sagittal seed. As a result, Dice and surface-distance curves remain comparatively stable across the organ, avoiding both the mid-range trough of 1-axial and the near-seed dip of 3-axial. Because all three settings use the same trained network and differ only in seed placement, the gap is attributable to seed geometry rather than training or annotation count.
\begin{figure}[!t]
\centering
\includegraphics[width=0.95\textwidth]{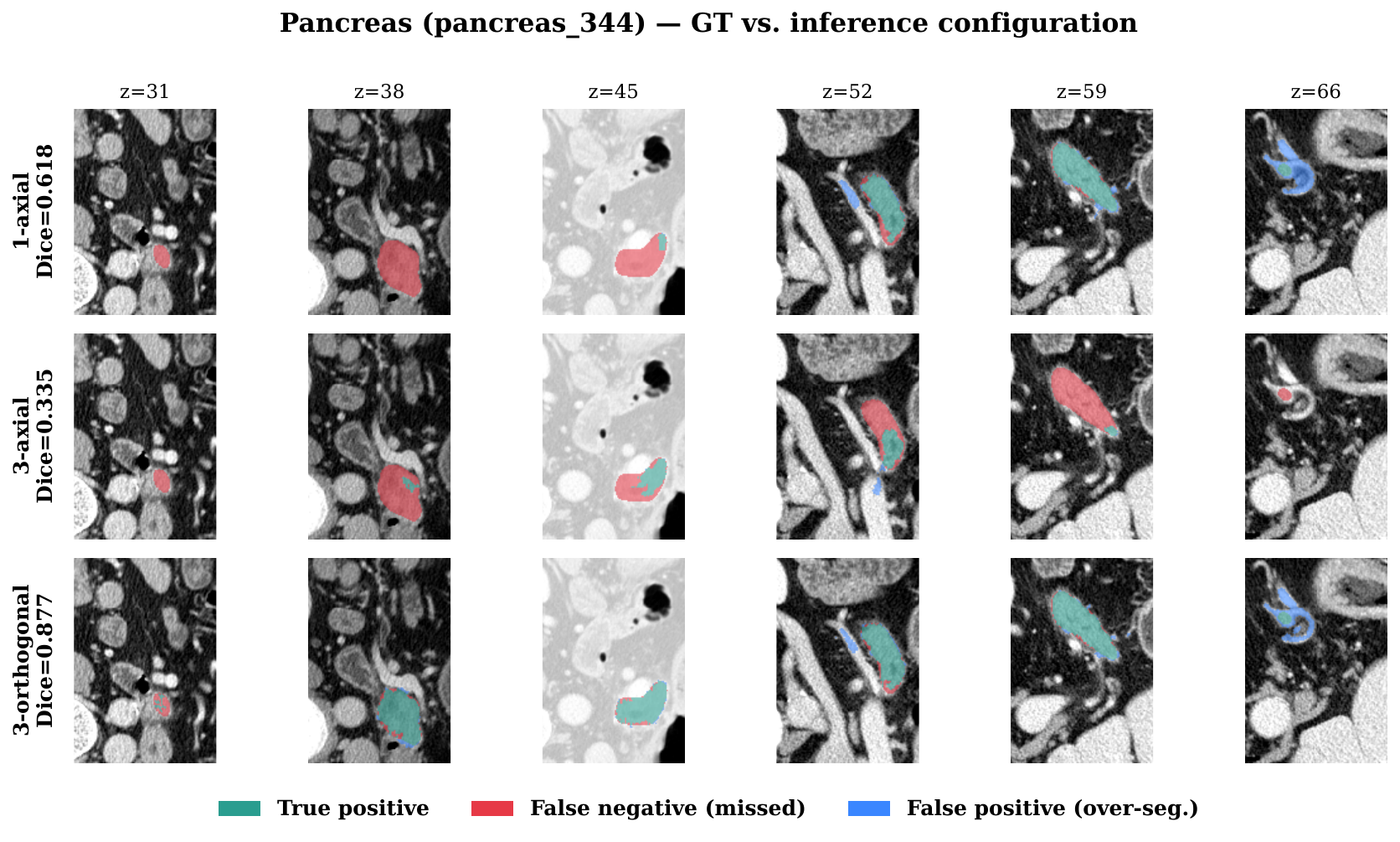}
\vspace{-0.5em}
\caption{Representative pancreas case on the triplanar Sli2Vol backbone. Green: correct; red: missed; blue: leakage. 1-axial and 3-axial miss the head and tail far from the axial seed; 3-orthogonal recovers the full organ. All rows share network weights and differ only in seed geometry.}
\label{fig:qual_pancreas}
\vspace{-1.5em}
\end{figure}

\textbf{Qualitative Results:} Figure~\ref{fig:qual_pancreas} shows the spatial pattern behind these numbers. The $1$-axial row is accurate on the slices near its seed ($z{=}52$, $z{=}59$) but misses the head and tail of the organ, far from the single axial seed, where whole sections go unsegmented ($z{=}38$, $z{=}45$). $3$-axial does not repair this: with all three seeds still in the axial plane, the same distant slices remain missed, and the extra coplanar chains add conflicting evidence across the mid-organ, giving the lowest Dice of the three ($0.335$). $3$-orthogonal instead recovers the organ along its full length, as the coronal and sagittal chains supply reliable predictions exactly where the axial chain has decayed, raising Dice to $0.877$. All three rows come from identical network weights and differ only in where the seeds are placed, so the difference is purely one of seed geometry, orthogonality rather than annotation count, the same conclusion the quantitative and distance-curve results reach.

\textbf{Assumptions and Limitations.}
Orthogonal seeding assumes a compact target with comparable extent across all three axes, so that off-axis seeds reduce effective propagation distance. Its benefit may therefore be smaller for thin, elongated, or branching structures. Our evaluation is limited to abdominal CT and three organs, liver, pancreas, and spleen, so transfer to other modalities, such as MRI, and to different anatomical geometries remains to be tested. The deformable backbone is trained only in the triplanar regime; therefore, the axial-only with orthogonal-seeding setting is absent for TransMorph, and our cross-paradigm claim is based on matched multi-axis training rather than an axial-only deformable baseline. Finally, our fusion rule weights each chain only by seed distance, which explains why the hard-vote variant required for the label-free deformable backbone can admit off-axis boundary noise. A confidence-aware, per-chain reliability estimate is a natural remedy and an important direction for future work.

\section{Conclusion}

We investigated what drives single-slice mask propagation quality and found that the key factor is inference-time seed geometry, not the registration backbone or its training. Propagating one axial, one coronal, and one sagittal seed with label-free distance-weighted fusion improves Dice by $21.9\%$, increases NSD by $25.5\%$, and reduces AHD by $53.5\%$ over the single-axis baseline, yielding better anatomy-aligned segmentations, especially at organ surfaces. A budget-matched three-axial control provides no benefit and can even degrade performance, showing that the gain comes from orthogonality rather than annotation count. These results indicate that axial-only propagation leaves many slices far from a reliable seed, while orthogonal seeds reduce effective propagation distance and provide complementary anatomical views. The same trend holds for correspondence-based and deformable backbones, making orthogonal seeding a plug-and-play inference strategy that improves accuracy without architectural changes, retraining, or dense voxel-level annotation. Although orthogonal seeding increases the manual annotation requirement from one slice to three, the resulting gains in segmentation accuracy and surface alignment justify this modest cost. Future work should explore stronger fusion strategies that further improve surface reconstruction without requiring more than three annotated slices.

\begin{credits}
\subsubsection{\ackname} 
The authors acknowledge the support of the Kahlert School of Computing and the Scientific Computing and Imaging Institute at the University of Utah.
\subsubsection{\discintname}
The authors have no competing interests to declare that are relevant to the content of this article.
\end{credits}
\bibliographystyle{splncs04}
\bibliography{mybibliography}

\end{document}